%% file: main.tex
\documentclass{article}

\usepackage{arxiv}
\usepackage[utf8]{inputenc}

\usepackage[numbers]{natbib}
\usepackage{hyperref}
\usepackage{graphicx}
\usepackage{amsmath,amssymb}
\usepackage{subfig}
\usepackage{algpseudocode}
\usepackage{algorithm}
\usepackage{amsthm}

\title{Logic-based Reward Shaping for Multi-Agent Reinforcement Learning}

\author{

   \href{https://scholar.google.com/citations?user=c5JdTqMAAAAJ} {\hspace{1mm}Ingy ElSayed-Aly}\\
	Department of Computer Science\\
	University of Virginia\\
	Charlottesville, VA 22903 \\
	\texttt{ie3ne@virginia.edu} \\
	
	\And
	{\hspace{1mm}Lu Feng}\\
	Department of Computer Science\\
	University of Virginia\\
	Charlottesville, VA 22903 \\
	\texttt{lu.feng@virginia.edu } \\
	
}

\renewcommand{\undertitle}{Preprint Technical Report}

\hypersetup{
pdftitle={Logic-based Reward Shaping for Multi-Agent Reinforcement Learning},
pdfauthor={Ingy ElSayed-Aly, Lu Feng},
pdfkeywords={Reward Shaping, Multi-Agent Reinforcement Learning, Formal Methods},
}

\begin{document}

\maketitle



\begin{abstract}
Reinforcement learning (RL) relies heavily on exploration to learn from its environment and maximize observed rewards. Therefore, it is essential to design a reward function that guarantees optimal learning from the received experience. Previous work has combined automata and logic based reward shaping with environment assumptions to provide an automatic mechanism to synthesize the reward function based on the task. However, there is limited work on how to expand logic-based reward shaping to Multi-Agent Reinforcement Learning (MARL).  The environment will need to consider the joint state in order to keep track of other agents if the task requires cooperation, thus suffering from the curse of dimensionality with respect to the number of agents. This project explores how logic-based reward shaping for MARL can be designed for different scenarios and tasks. We present a novel method for  semi-centralized logic-based MARL reward shaping that is scalable in the number of agents and evaluate it in multiple scenarios.
\end{abstract}

\section{Introduction}

 With the rise of cyber-physical systems (CPS) and the age of the Internet of Things (IoT) we will have more multi-agent systems to control using advanced techniques ~\cite{li2015internet}. Some systems will be given one same task to execute whereas some will need to split into subgroups to execute several tasks. Moreover, the use of RL to train robots and devices is becoming more and more prevalent ~\cite{alur2015principles}. One of the main appeals of RL is its capability to adapt and learn in an unknown or partially known environment. However, this is only possible if the reward function is designed correctly for the problem the agent is attempting to solve.  
 
 Designing the reward function is still mostly done manually in most cases and is often hard to interpret ~\cite{li2019formal, garcia2015comprehensive}. For this reason, logic-based reward shaping is a promising technique which uses the flexibility of Linear Temporal Logic (LTL) to automate the reward function design process based on the task.  LTL has also proven to be an expressive and widely used task specification language in robotic applications as well as a possible candidate for translation from natural language~\cite{feng2011automated}. Furthermore, we believe that this could be very useful in the multi-agent case where it becomes even more tricky to design the reward function. In MARL, agents' interactions can affect each other’s rewards and make it harder to train.  In addition, the
joint state space and joint action space increase exponentially with
the number of agents which poses scalability issues.

 Using logic-based reward shaping with single agent RL has demonstrated efficient task learning as well as flexibility of expression of the task thanks to LTL ~\cite{baier2008principles}. Once a Limit-Deterministic B\"uchi Automaton (LDBA) is synthesized based on the LTL task, we can obtain a product Markov Decision Process (MDP) based on the environment MDP and the LDBA automaton. Then, we can use the accepting states to reward the agent such that it visits the accepting states infinitely often ~\cite{hasanbeig2020cautious, bozkurt2020control}. With an automated method to synthesize the product MDP, the reward function only depends on the product MDP state. 
 
To the best of our knowledge there is only one previous work attempting to use logic-based reward shaping methods in Multi-Agent Systems, which is based on a more limited fragment of LTL and addresses only tasks that can be fully decomposed into individual tasks ~\cite{neary2020reward}. Furthermore, MARL is notoriously difficult to train because of the interactions between agents and it is possible that the logic-based reward shaping methods that exist will not work in this setting ~\cite{bucsoniu2010multi}. 

In this report, we present a semi-centralized approach for reward shaping in multi-agent systems, where we synthesize a centralized LDBA which monitors the agents' progress with respect to a given LTL formula. When a set of observed labels violate the LDBA specification, the automaton transitions to a trap state and returns a large negative reward. If the set of observed labels or selected epsilon-actions transition the LDBA to an accepting state, the agents receives a positive reward. We follow the joint rewards mechanism for this approach; that is, all agents receive the same rewards based on the common LDBA state.
 
This report addresses the need for an automated framework converting tasks to reward functions for their respective agents or a team of agents. Inspired by the logic-based reward shaping mechanisms existing for single agent RL, we develop a novel framework for logic-based reward shaping in MARL. Our contributions include:
\begin{itemize}
    \item  We develop a semi-centralized approach for multi-agent reward shaping that is scalable in the number of agents. 
    \item We showcase our approach via experimental evaluation on multiple benchmarks problems.
\end{itemize}

This report is structured as follows. We first discuss relevant prior work in section \ref{related work}, then we present important background knowledge with respect to reward shaping in RL in section \ref{background}. We identify a motivating example and problem statement in section \ref{motivating}. In section \ref{approach} , we explain our proposed method in section. We proceed to evaluate our method and present the relevant results in section \ref{experiments}.  Finally,  we discuss limitations in section \ref{discussion} before concluding in section \ref{conclusion}.

\section{Related Work} 
\label{related work}

\paragraph{Logic-based Reward Shaping.} Linear Temporal Logic (LTL) is a commonly used specification language in formal methods for safety-critical systems ~\cite{alur2015principles,baier2008principles}. For example, LTL has been used to express complex task specifications for robotic planning and control~\cite{kress2009temporal,ulusoy2013optimality}.  Tasks expressed in LTL can be represented as a Limit-Deterministic B\"uchi Automaton (LDBA) clearly highlighting accepting states ~\cite{hasanbeig2020cautious}. The product of the obtained LDBA and the environment MDP can be used to designed a logic based reward function. In Hasanbeig et al. ~\cite{hasanbeig2020cautious} the reward function is purely based on visiting the accepting states whereas in Bozkurt et al. \cite{bozkurt2020control} the reward function is based on the path that is visited. In Kuo et al. ~\cite{kuo2020encoding} and in Elbarbari et al. ~\cite{elbarbari2021ltlf}, the authors explore different ways to capture metrics of progress within the logic synthesized automaton in order to help guide the learning more effectively.  

Logic based reward shaping can also be used in continuous control use cases ~\cite{cheng2019end, li2019formal}. In the context of RL, logic-based reward shaping has often been augmented by using control barrier functions to address probabilistic safety guarantees ~\cite{hasanbeig2020cautious,li2019formal,cheng2019end}. In the context of multi-agent systems, there are ways to plan using LTL without learning~\cite{guo2014cooperative, kress2009temporal}. There is one recent work considering safety in MARL by modifying the rewards the agents receives given the output of a logic-synthesized shield ~\cite{elsayed2021safe}. However, this work cannot be considered reward shaping because it does not guide the agent towards desired accepting states. 

Some works consider the co-safe fragment of LTL and instead convert the tasks to Mealy Machines to construct so-called Reward Machines (RM)~\cite{camacho2019ltl}. The work on RMs has recently been expanded to include cooperative MARL through task decomposition to individual agent tasks and decentralized Q-learning ~\cite{neary2020reward}. Because of the use of the RMs, the method is limited to the co-safe fragment of LTL in addition to considering only tasks that can be fully decentralized. This method also needs to decompose the LTL tasks in order for each agent to learn one specific task.


\paragraph{Multi-agent Reinforcement Learning} In Reinforcement Learning an agent learns an optimal behavior based on multiple trials. The agent interacts with an unknown environment usually modeled as an MDP. At each step, the agent chooses some action $a$, the environment moves it to the next state based on a transition probability and the agent receives a reward $ r $. The aim of the agent is to adapt its behavior in order to maximize the expected return $ \mathcal{R}  = \Sigma_{t = 0}^{\inf} \gamma^t r_t $ where $0 < \gamma \leq 1$ is a discount factor ~\cite{nowe2012game}.

In multi-agent reinforcement learning (MARL), at each time step $t$, each agent selects an action simultaneously, this set of all agent actions is called the joint action. The environment will then return a reward for each agent. In MARL, the reward function may favor competition or cooperation~\cite{bucsoniu2010multi} MARL algorithms can be considered Joint Action learners or Independent Learners~\cite{bucsoniu2010multi,nowe2012game}. For Joint Action learning algorithms, each agent considers all other agents, leading to complete communication but also poor scalability. In practice, it is unlikely that agents will need to communicate and share information with all other agents at each step. Independent Learners consider other agents when they ``detect" a need to coordinate or simply consider other agents part of the environment such in Independent Q-learning ~\cite{watkins1989learning}. Interestingly, despite Q-functions learned in Independent Q-Learning relying only on one agent's awareness, Independent Q-Learning has successfully been applied in multiple multi-agent settings with some limitations ~\cite{bucsoniu2010multi,tuyls2006evolutionary}. In the case of a larger state space, deep learning has proven to be a useful tool to generalize across states and reduces the need for manual feature design  ~\cite{hernandez2019survey,lowe2017multi}.


\section{Background}
\label{background}


\begin{figure}
\centering
\subfloat [a][MDP of an example environment]{
  \includegraphics[width=.4\linewidth]{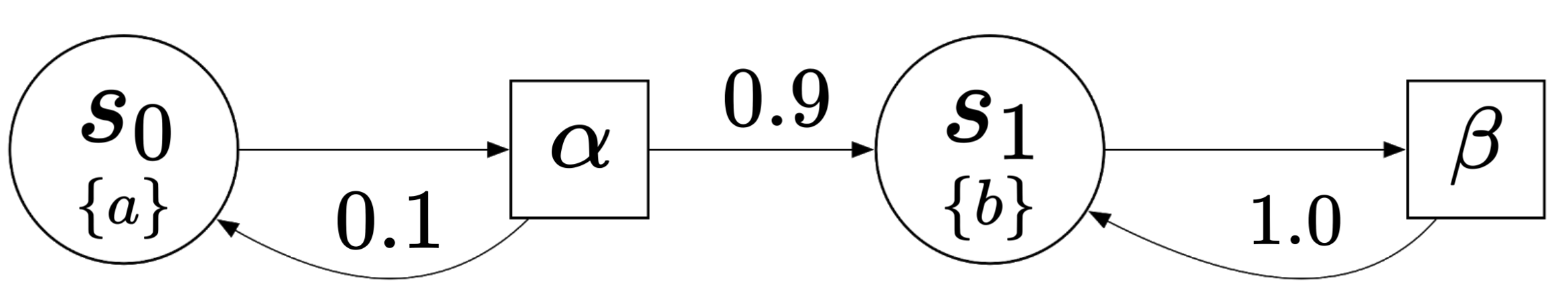}
  \label{fig:mdp}
}
\subfloat [b][LDBA for LTL formula $\Diamond \Box a \vee \Diamond \Box b$]{
  \includegraphics[width=.35\linewidth]{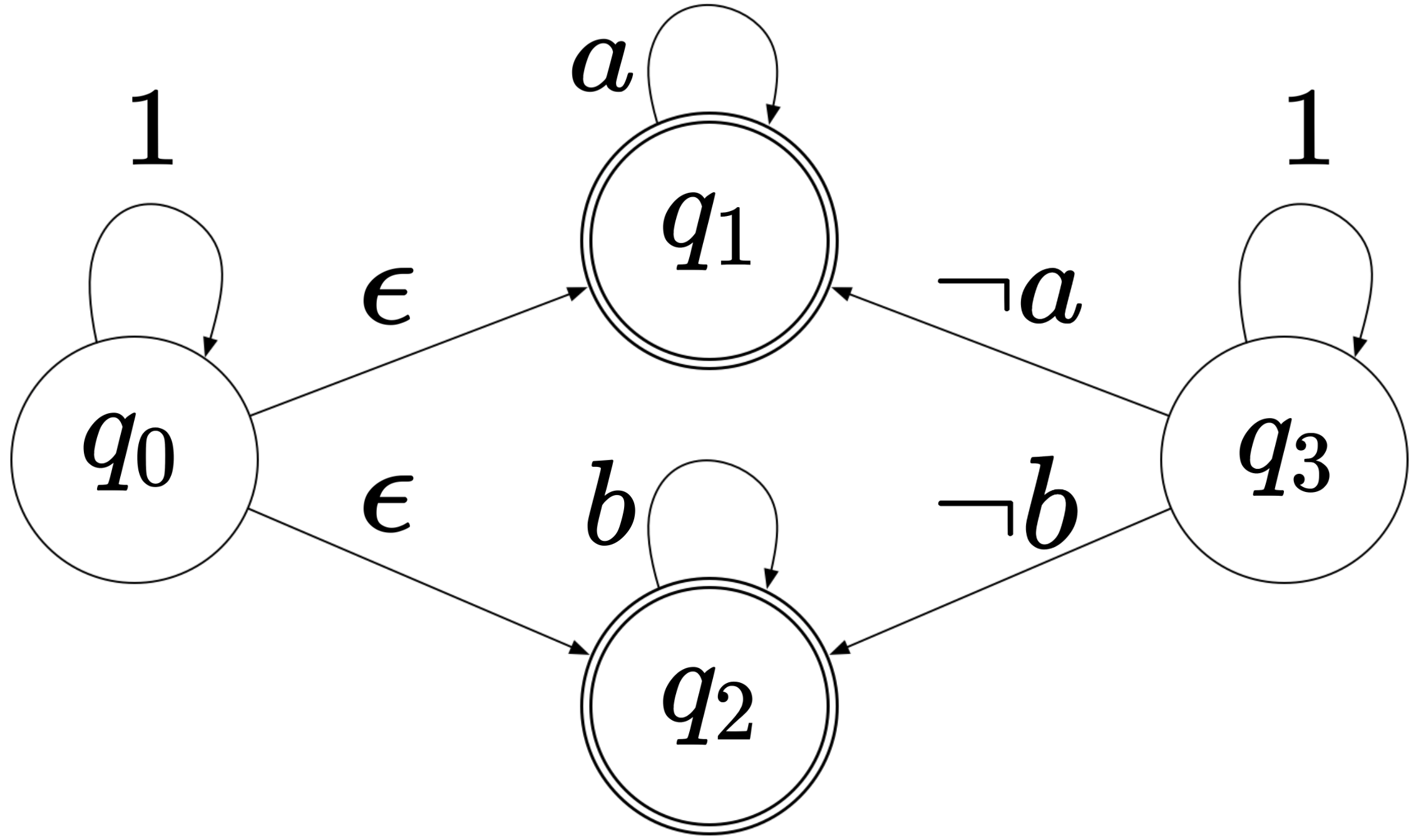}
  \label{fig:ldba}
}%
\\
\subfloat [c][Resulting product MDP]{
  \includegraphics[width=.65\linewidth]{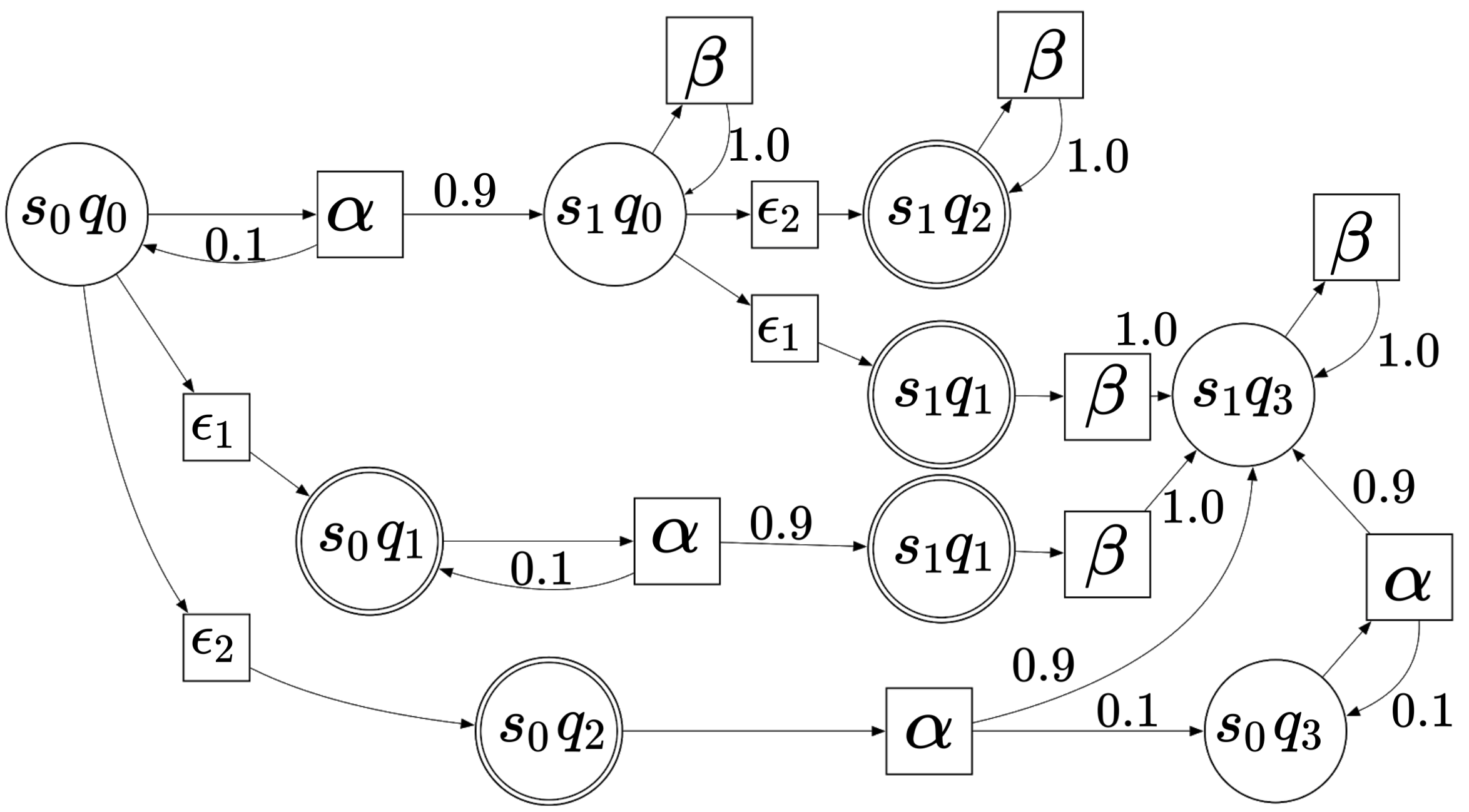}
  \label{fig:product}
}
\caption{Constructing the Logic-based Structure based on the Environment}
\label{fig:bigfig}
\end{figure}

In this section, we will introduce the important notions of logic-based reward shaping based on an example adapted from~\cite{hasanbeig2020cautious,bozkurt2020control}.

\paragraph{Markov Decision Process (MDP).} In figure \ref{fig:mdp}, is an example taken from ~\cite{hasanbeig2020cautious} Markov Decision Process (MDP) which is a structure often used to abstract the environment in reinforcement learning. One of the important features about MDPs is that it can represent probabilistic environments such as our example in which action $\alpha$ taken in state $s_0$ leads to state $s_1$ with probability $0.9$ and leads back to $s_0$ with probability $0.1$. States in MDPs can also have labels such as $a$ for $s_0$ and  $b$ for $s_1$. 

\paragraph{Limit Deterministic B\"uchi Automaton (LDBA).} In figure \ref{fig:ldba}, we see the representation of a Limit Deterministic B\"uchi Automaton (LDBA) which can be automatically generated based on an LTL formula taken from~\cite{bozkurt2020control}. The main features to notice about LDBA are that there are accepting states $\{q_1, q_2\}$ and that the accepting states and the starting states are two disjoint sets separated by non-deterministic $\epsilon$-transitions~\cite{bozkurt2020control}. In this figure, the formula is equivalent to the English interpretation: ``Eventually Always a Or Eventually Always b". This means that eventually we must only find ``a" consistently or only ``b" consistently. Here ``a" and ``b" refer to labels of certain states in the MDP, they could be alternate goal locations. Thus, the formula would be encouraging the agent to find one of the goal locations and stay there. Therefore the only accepting states are $\{q_1, q_2\}$, since once you find a ``b" you have to stick to that label otherwise you violate the violate the policy and same goes for $q_1$ with ``a". Here $q_3$ is a failure state from which you cannot recover because of the nature of the specification.

\paragraph{Product MDP.} Figure \ref{fig:product} we show how the LDBA from figure \ref{fig:ldba} and the MDP from figure \ref{fig:mdp} can be combined into a product MDP as formally defined in~\cite{bozkurt2020control}. Each state of the new MDP will correspond to a state in the LDBA and a state of the MDP based on the transitions that have been taken and the labels that have been encountered. Moreover, each $\epsilon$-transition in the LDBA gets translated into an action that does not change the MDP state in the product MDP, only the LDBA state. From state $s_0q_0$ the possible actions are $\alpha$ or the $\epsilon$-transitions (because of $q_0$). In this MDP, the accepting states are $\{s_0q_1, s_0q_2, s_1q_1, s_1q_2\}$ and the states $\{s_0q_3, s_1q_3\}$ form a non-accepting sink component. In the context of learning, the $\epsilon$-actions can be seen as a guess which may or may not be taken at the right time, in which case they may lead to non-accepting sink component. This is the case if action $\epsilon_2$ is taken at state $s_0q_0$. However, in the following episodes, the agent should learn that taking action $\epsilon_2$ from state $s_1q_0$ leads to an accepting strong connected component (SCC).  \\

\paragraph{Logic based reward shaping} The main idea in Bozkurt et al. ~\cite{bozkurt2020control} is to encourage the RL agent to learn a control policy that maximizes the probability of satisfying the B\"uchi condition for the LTL specification in an arbitrary MDP. The B\"uchi condition is satisfied when the accepting states of the LDBA automaton are visited infinitely often by the agent's policy. They optimize for this behavior by defining the reward function such that the value of each state in the set of accepting states approaches $1$. If an agent visits an accepting state, the agent receives a reward $(1-\gamma_B)$ and uses an accepting state specific discount value $\gamma_B$. If the agents visits a non-accepting state, the agent receives a reward of $0$ with a discount value of $\gamma$. The probability of satisfying the B\"uchi condition is maximized as the discount factor $\gamma$ approaches $1^-$. In other work, encouraging the agent to visit the accepting states infinitely is translated into different types of reward functions ~\cite{hasanbeig2020cautious,kuo2020encoding}. In some cases the reward function completely replaces the environment reward and in others balancing factor are used to quantify the relative weight of the different sources of rewards ~\cite{li2019formal, hasanbeig2020cautious}. 







\section{Motivating example}
\label{motivating}
\begin{figure}
\centering
\subfloat [c][Grid representation of Motivating example MDP]{
  \includegraphics[width=.2\linewidth]{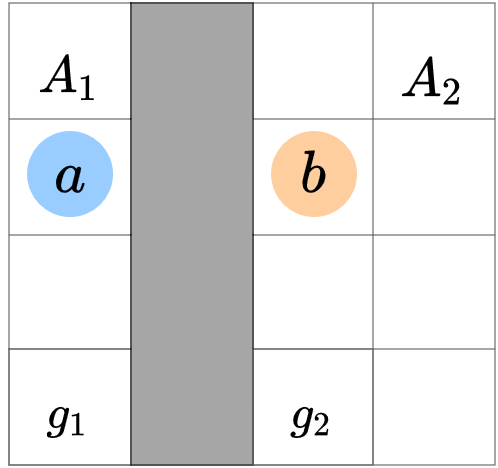}
  \label{fig:example}
}
\\
\subfloat [a][Automaton for agent $A_1$,\\ $\phi_1= (\neg a \cup b) \land \Diamond \square g_1 $]{
  \includegraphics[width=.35\linewidth]{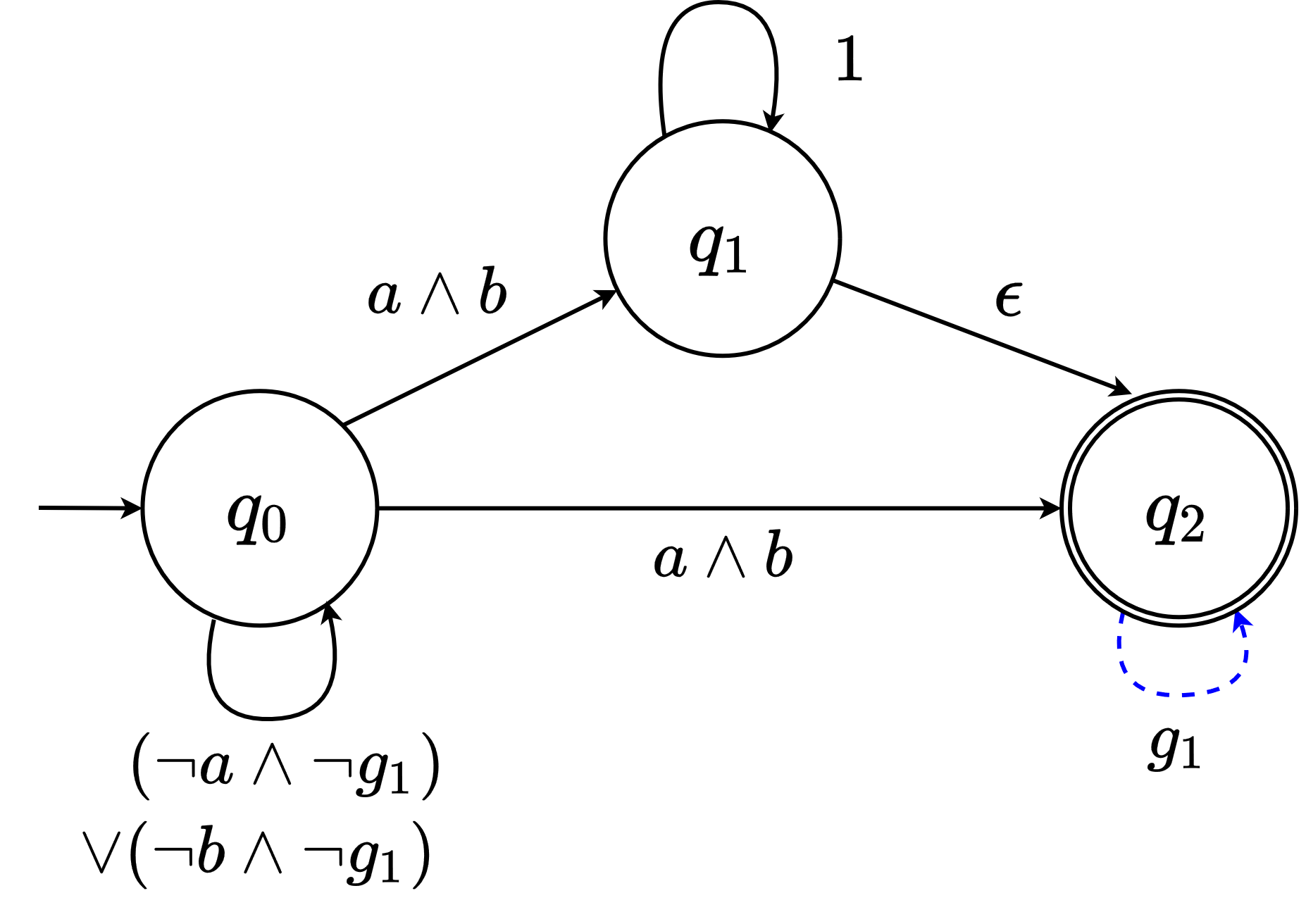}
  \label{fig:example_a1}
}
\subfloat [b][Automaton for agent $A_2$,\\ $ \phi_2= (\neg b \cup a) \land \Diamond \square g_2 $] {
  \includegraphics[width=.35\linewidth]{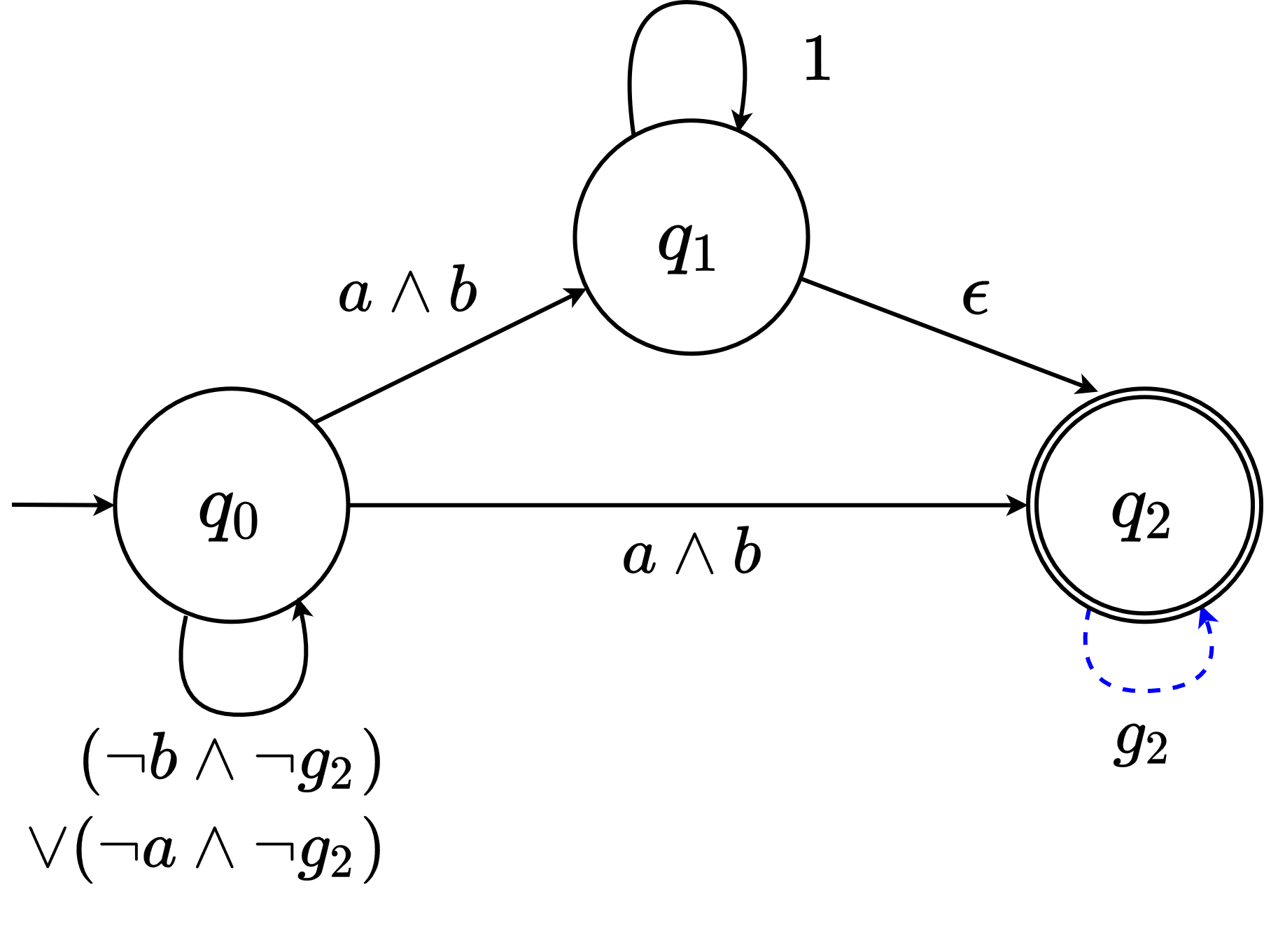}
  \label{fig:example_a2}
}%

\caption{Motivating Example}
\label{fig:motivatingex}
\end{figure}

In this work, we first introduce a two agent scenario adapted from an example presented in Neary et al. ~\cite{neary2020reward} to motivate our approach. In figure \ref{fig:motivatingex},  agent $A_1$ needs agent $A_2$ to press button $b$ before proceeding towards its goal $g_1$ and vice versa. Effectively, both agents need to be pressing the button at the same time before they can reach their goals. In this example, the environment does not enforce this constraint, only that the agents may not pass the wall delineated in gray. 

The example in figure \ref{fig:motivatingex} illustrates the kind of problem which cannot be resolved with a simple independent learners strategy because an agent needs to be aware of labels triggered by other agents.  In this report, we demonstrate that this can be solved without explicitly splitting the task into multiple individual sub-tasks and use the MARL to learn the sub-tasks. 

We consider the problem of learning agent policies in a stochastic multi-agent environment satisfying a desired specification. The environment is modeled as an MDP which may be an abstraction of the actual environment but in which should be completely characterized for each agent, i.e. one agent's action and state will not affect the other agents actions and  MDP states. The desired objective is given by an LTL formula which specifies all requirements to be met and if appropriate in which order. Our goal is for agents to learn policies such that all tasks and requirements are satisfied independently of the underlying MARL algorithm $\mathbb{A}$.

\section{Approach: Semi-centralized Reward Shaping}
\label{approach}

\begin{figure}
\centering
\includegraphics[scale=0.3]{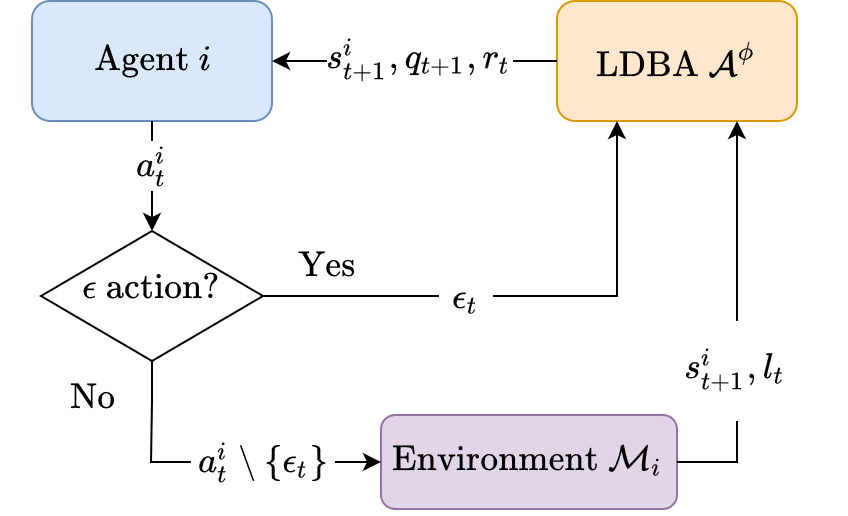}
\caption{Semi-centralized Reward Shaping}
\label{fig:diagram}
\end{figure}

In this section, we introduce a \textit{centralized automaton} $\mathcal{A}^\phi$ in the MARL learning process. In the following, we explain how our method can be used to solve problems requiring agent coordination. 

Figure \ref{fig:diagram} illustrates the interaction between each agent, the LDBA $\mathcal{A}^\phi$ synthesized from the LTL specification $\phi$ and the environment represented as an MDP $\mathcal{M}$ from each agent's point of view. At time step $t$,  agent $i$ chooses some action $a_t^i$ based on the underlying algorithm $\mathbb{A}$ . Then if the action is an $\epsilon$-action, the epsilon action is forwarded to $\mathcal{A}^\phi$ to update the automaton state. If it is not an $\epsilon$-action then it is sent along with the actions of the other agents to the environment. From the environment, we receive the new set of states and a set of labels. The labeling function need not be part of the environment but can be determined as part of the current states, actions and new states. The labels allow us to transition to the next automaton states (i.e. $q_{t+1}$) in the LDBA $\mathcal{A}^\phi$ for non $\epsilon$-transitions. In this method, the agents' states are augmented with the automaton state $q_{t}$ at time t and $q_t$ is identical for each agent. In the case of decentralized execution, each agent may have a copy of the LDBA $\mathcal{A}^\phi$ for which the state $q_t$ is synced at every time step. The purpose of the automaton in this approach is to identify the different states of progress for an LTL specification which can happen regardless of the number of agents involved while giving agents flexibility in the way that the tasks are assigned and carried out. Without syncing the automaton states with the other agents, the product MDP for any one agent is incomplete and cannot reach the accepting states.

\input{algorithm1}

In algorithm \ref{algorithm1}, we demonstrate how to implement this method with a generic MARL algorithm $\mathbb{A}$. Instead of explicitly building product a MDP for each agent, we can track and synchronize the progress of each agent in the LDBA. We remark that an agent's state is an augmented state composed of the original MDP state and the automaton state. We first choose an action for each agent $i$ from the available actions at state $s^i_t$ which include $\epsilon$ actions if $\epsilon$ actions are available at state $q_t$ of the automaton. The environment here represented by MDP $\mathcal{M}_i$ from the point of view of an agent $i$ transitions based on the selected non-$\epsilon$ actions. The labels for the new state are then retrieved to transition the automaton for the label-based transitions. For regular transitions, the automaton progresses based on a joint set of labels $l_t$ which corresponds to the union of labels observed by all the agents at time $t$. For $\epsilon$-action based transitions, only one $\epsilon$-action is allowed per time step in addition to label based transitions where $\epsilon$-actions are resolved before label-based transitions. This is consistent with the definition of an LDBA for which the $\epsilon$-moves can be taken at any time as long as they are available at the current state of $\mathcal{A}^\phi$ ~\cite{bozkurt2020control}. Finally, the MARL algorithm $\mathbb{A}$ is updated with the augmented state and reward received from reward shaping. Using one LDBA  in combination with not explicitly building a product MDP enables us to sidestep the issue of exponential increase in the number of product MDP states proportional to the number of agents $n$ . 

\begin{figure}
\centering
\includegraphics[scale=0.11]{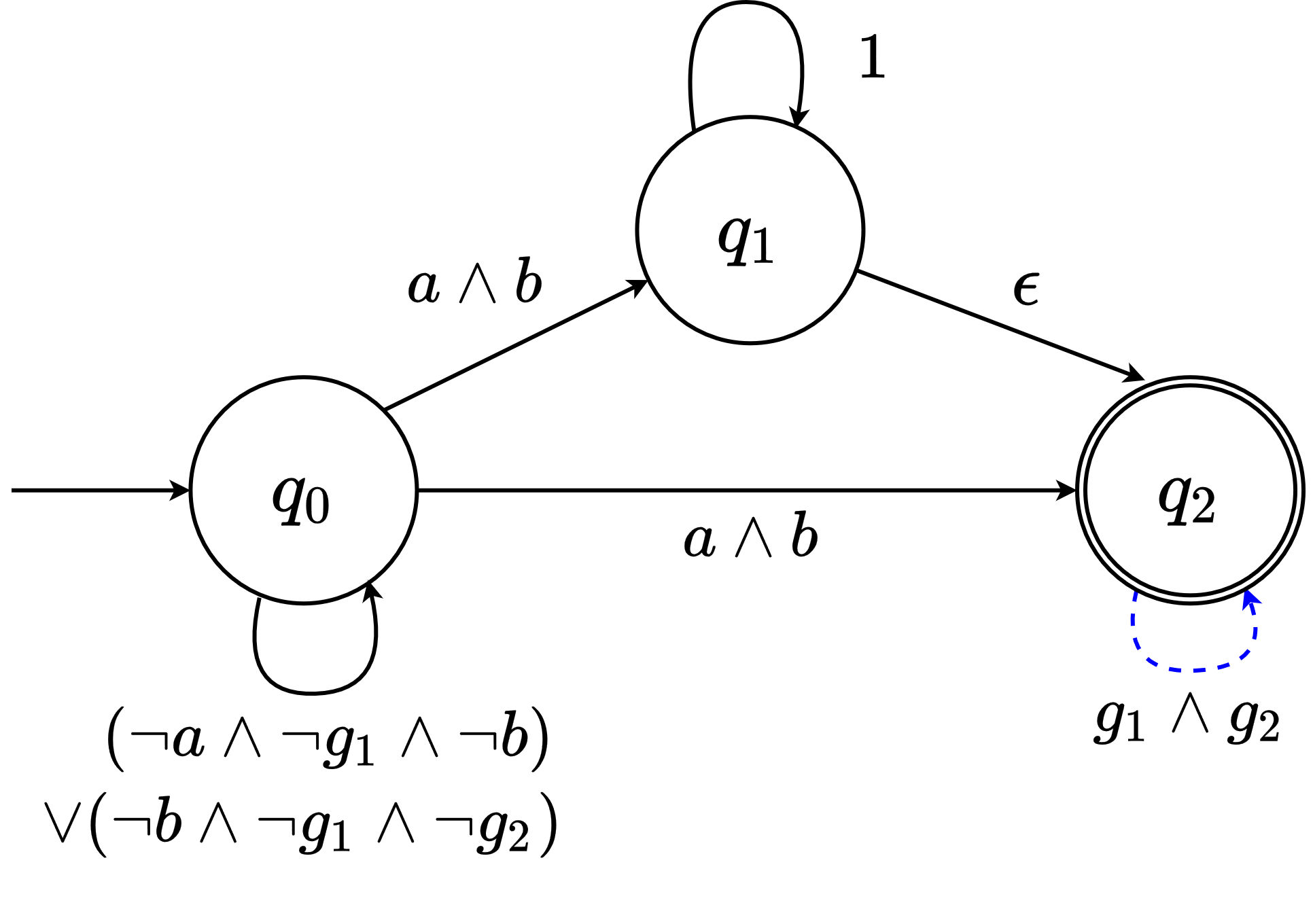}
\caption{Automaton Representing Combined LTL Tasks}
\label{fig:automatonf3}
\end{figure}

In the motivating example (fig:\ref{fig:motivatingex}), the first agent's automaton would check for button $b$ is triggered to be able to pass towards goal $g_1$. The second agent's automaton would include button $a$ that can only be reached by the first agent. The respective LTL formulas can be expressed as  $\phi_1= (\neg a \cup b) \land \Diamond \square g1 $ and $ \phi_2= (\neg b \cup a) \land \Diamond \square g_2 $.  In the case of the running example from fig.\ref{fig:motivatingex}, we can now express the tasks as follows : 
\begin{align*}
\phi_3 = &((\Diamond a \land \neg g_1) \cup (a \land b)) \\
		\land & ((\Diamond b \land \neg g_2) \cup (a\land b)) \\
        \land & (\Diamond\square g_1) \land (\Diamond\square g_2)
\end{align*}
The first line ensures that agents do not attempt to go to $g_1$ without the buttons being cleared, the second ensures that the agents do not attempt to go to $g_2$ without the buttons being triggered and the last last line encourages the agents to eventually find the goals $g_1$ and $g_2$. The resulting automaton is displayed in fig.~\ref{fig:automatonf3}. Notice that the transition from state $q_1$ to state $q_2$ requires an epsilon-action. This example demonstrates how using a common automaton can give us flexibility in the the task assignment. Here the user does not need to know which agent should carry out which tasks. The user can still prevent the agents from doing tasks out of order and effectively synchronize the agents. If the buttons represent doorbells and the goals represent opening the doors, the simulation environment could allow the agent to opening the doors without ringing the doorbells which can be an undesirable behavior. Furthermore, this example shows that a combined specification $\phi_3$ does not necessarily result in a large increase in the number of automaton states (fig.\ref{fig:automatonf3} ). 
\paragraph{Complexity.} In terms of time complexity, the main added complexity is the synthesis of the LDBA $\mathcal{A}^\phi$ which can be computed once at the beginning and the automaton sent to all agents. In terms of space complexity, we avoid exponential complexity in the number of agents by considering the point of view of only one agent at a time. However, instead of the memory required by agent $i$ being a product of $\mathcal{M}_i$'s state space and $\mathcal{M}_i$'s action space, we now have both an augmented state space and an augmented action space for each agent. However, we argue that although more complex tasks may require a larger number of  automaton states, the LDBA of the joint specification may exploit symmetry within the tasks (fig.~\ref{fig:automatonf3}). In terms of communication, our method only requires that the value of the automaton state and selection of $\epsilon$-actions be transmitted which is why our method can be considered semi-centralized.  
\paragraph{Correctness.} We show that the reward function obtained is correct with respect to the joint LTL specification $\phi$. Using an LDBA allows us to identify all the different traces through which the joint LTL specification for the MARL tasks may be satisfied. Thus, assuming the labels are identified correctly, following the reward function obtained from the corresponding LDBA is guaranteed to be an accurate representation of the agents' progress. In other words, if the agents receive a non-zero reward then it is guaranteed to correspond to an accepting transition of $\mathcal{M}^x$. We further show that our method is equivalent to building a product MDP  $\mathcal{M}^x = \left(\mathcal{A}^\phi \times \mathcal{M}_0 \times \ldots \times \mathcal{M}_n\right)$, $\forall \text{ agents } \in n$ after resolving the $\epsilon$-transitions. In our algorithm, we implicitly build the product pairs $\left(\mathcal{A}^\phi \times \mathcal{M}_i\right), \forall i \in n$ where each agent $i$ is responsible for one pairing and the automaton $\mathcal{A}^\phi$ is synchronized. 

By definition of a pair product MDP, the actions allowed at a specific state are given by union of the actions allowed in the MDP state and the $\epsilon$-actions allowed in the automaton state. So if we consider the larger picture with all the agents, then the allowed actions at any given state of the product MDP is the union of the action space all pairs for that state which corresponds to the union of the automaton actions with the MDP action space of all agent MDPs. In our case here, the $\epsilon$-actions are not repeated since in the case of multiple $\epsilon$ -action only one is selected. 

The state space of the pair product MDPs is defined by all the possible combinations of the state space of $\mathcal{A}^\phi$ and the the state space of $\mathcal{M}_i$. If we extend that to all agents, the joint state space is characterized by $\left(\mathcal{A}^\phi \times \mathcal{M}_0\right) \times \ldots \times \left(\mathcal{A}^\phi \times \mathcal{M}_n\right)$. However, since the LDBA state across all agent pair product MDP is the same, this reduces to $\left(\mathcal{A}^\phi \times \mathcal{M}_0 \times \ldots \times \mathcal{M}_n\right)$, proving it is identical to the product MDP $\mathcal{M}^x$ state space.

The transition function of the product MDP $\mathcal{M}^x$ is defined by the product of the automaton transitions and the successive MDP transitions where if labels resulting in an automaton state transition are identified; the transition of the automaton and the transition of the MDP state can be combined (fig.~\ref{fig:bigfig}). Because each MDP $\mathcal{M}_i$ transition is fully characterized by one agent, the order of the agent environment transitions during a time step $t$ does not matter (allowing us to take them simultaneously). Therefore, since the set of observed labels $l_t$ at time $t$ is identical for all agents and only one common $\epsilon$ action can be taken, the transition computed by each agent for the automaton is deterministic and unique. Thus, the product MDP $\mathcal{M}^x$ transition function reduces to the pair product MDP transitions. 

This proves that all aspects of the product MDP $\mathcal{M}^x$ are consistent with our implementation in algorithm $\ref{algorithm1}$ of the pair product MDPs for each agent. 

\paragraph{Impact on Learning Performance.} This method is agnostic to the choice of a MARL Algorithm because the reward shaping interacts with the learner only via inputs and outputs, and does not rely on the inner-workings of the learning algorithm $\mathbb{A}$. The convergence of this method may depend on the underlying algorithm. However, there is a lack of theoretical convergence guarantees for MARL algorithms in general. Thus, we focus on showing empirical convergence in our experiments in Section \ref{experiments}. 

Because of the flexibility in the expression of an LTL specification, slightly different ways of specifying the same general tasks may lead to a different number of accepting transitions. The number of accepting transitions directly impacts the learning performance by providing more or less guidance with respect to progress in the LDBA. Note for example that the specifications for the motivating example ($\phi_3$) and the rendez-vous benchmark here denoted by $\phi_3'$ (presented in section \ref{experiments}) have similar tasks which consist in having both agents synchronize at locations $a$ and $b$ then proceeds to the goal locations $g_1$ and $g_2$. However, the specifications differ with $\phi_3' = \Diamond( (a\land b) \land \bigcirc\Diamond ( (g_1 \lor \bigcirc\Diamond g_1) \land ( g_2 \lor \bigcirc\Diamond g_2)) )$ resulting in the automaton shown in figure \ref{fig:automaton_spot_3} in Appendix \ref{spot_automata}.  Further analysis of both specifications shows that the winning region $w_3$ of $\phi_3'$ is covered by the winning region $w_r$ for $\phi_3'$; formally, $w_3 \subseteq w_r$. We show that in our experiments, a larger number of accepting transitions (i.e. more possible rewards during an accepting run) translates into better learning performance (fig.~\ref{fig:bench1_learning_curve}). 


\section{Experiments}
\label{experiments}
\begin{figure}
\centering
\subfloat [\emph{a}][Representation of the Environment MDP]{
  \raisebox{31 pt}{\includegraphics[width=.3\linewidth]{figures/motivating_ex.png}}
  \label{fig:bench1_mdp}
}
\subfloat [\emph{b}][Normalized Mean Return Learning Curve]{
  \includegraphics[width=.55\linewidth]{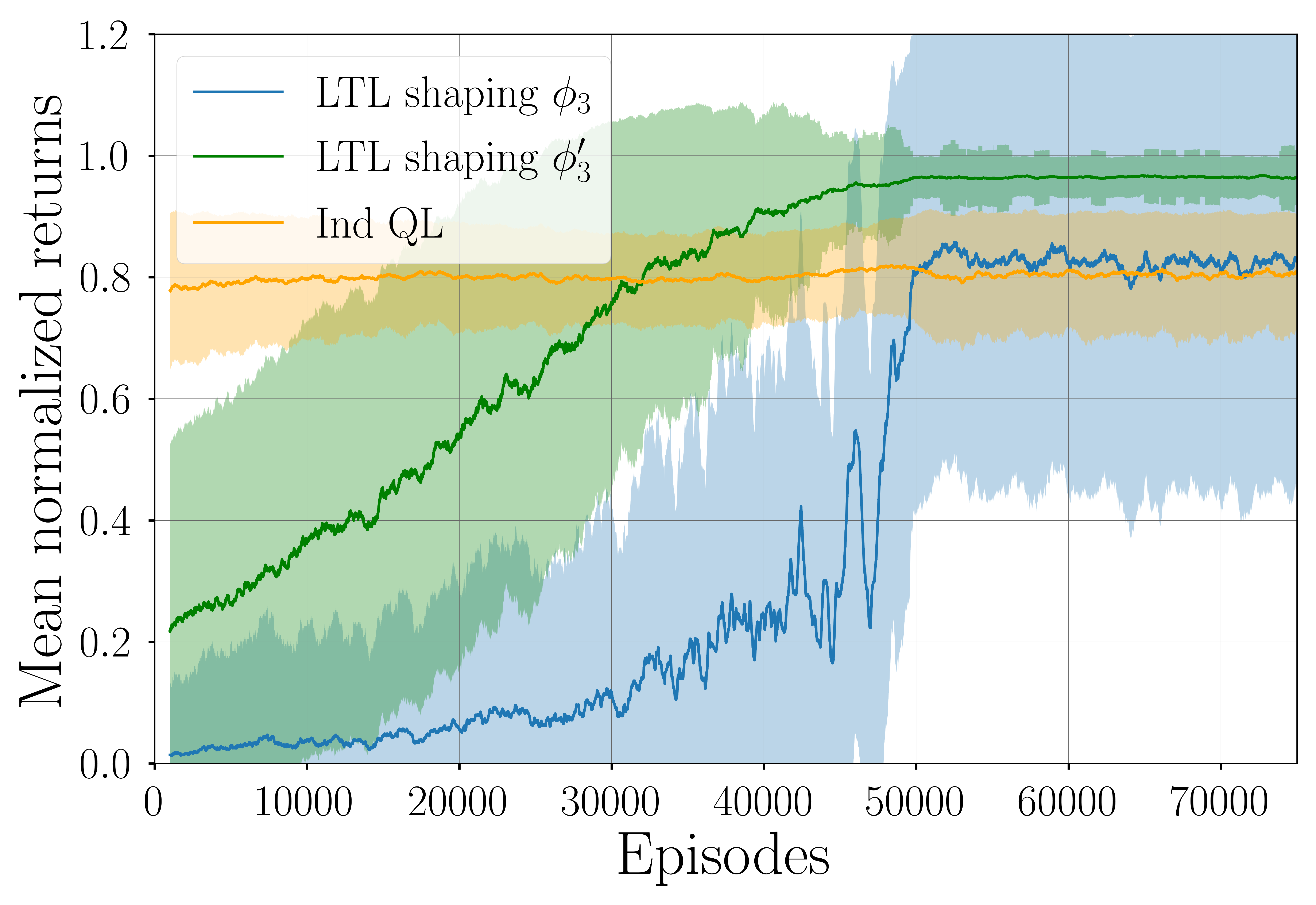}
  \label{fig:bench1_learning_curve}
}%
\caption{Results for the Motivating Example}
\label{fig:bench1}
\end{figure}

\begin{figure}
\centering
\subfloat [\emph{a}][Representation of the Environment MDP]{
  \raisebox{52 pt}{\includegraphics[width=.35\linewidth]{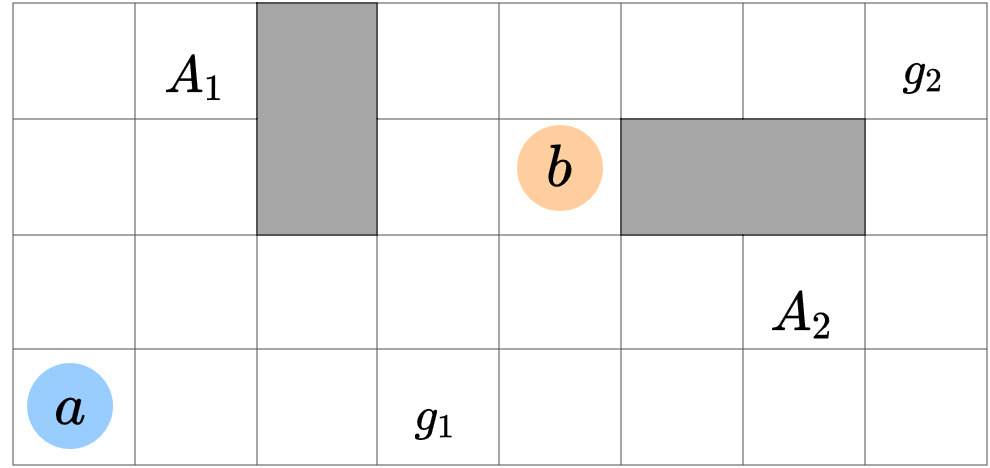}}
  \label{fig:bench2_mdp}
}
\subfloat [\emph{b}][Normalized Mean Return Learning Curve]{
  \includegraphics[width=.55\linewidth]{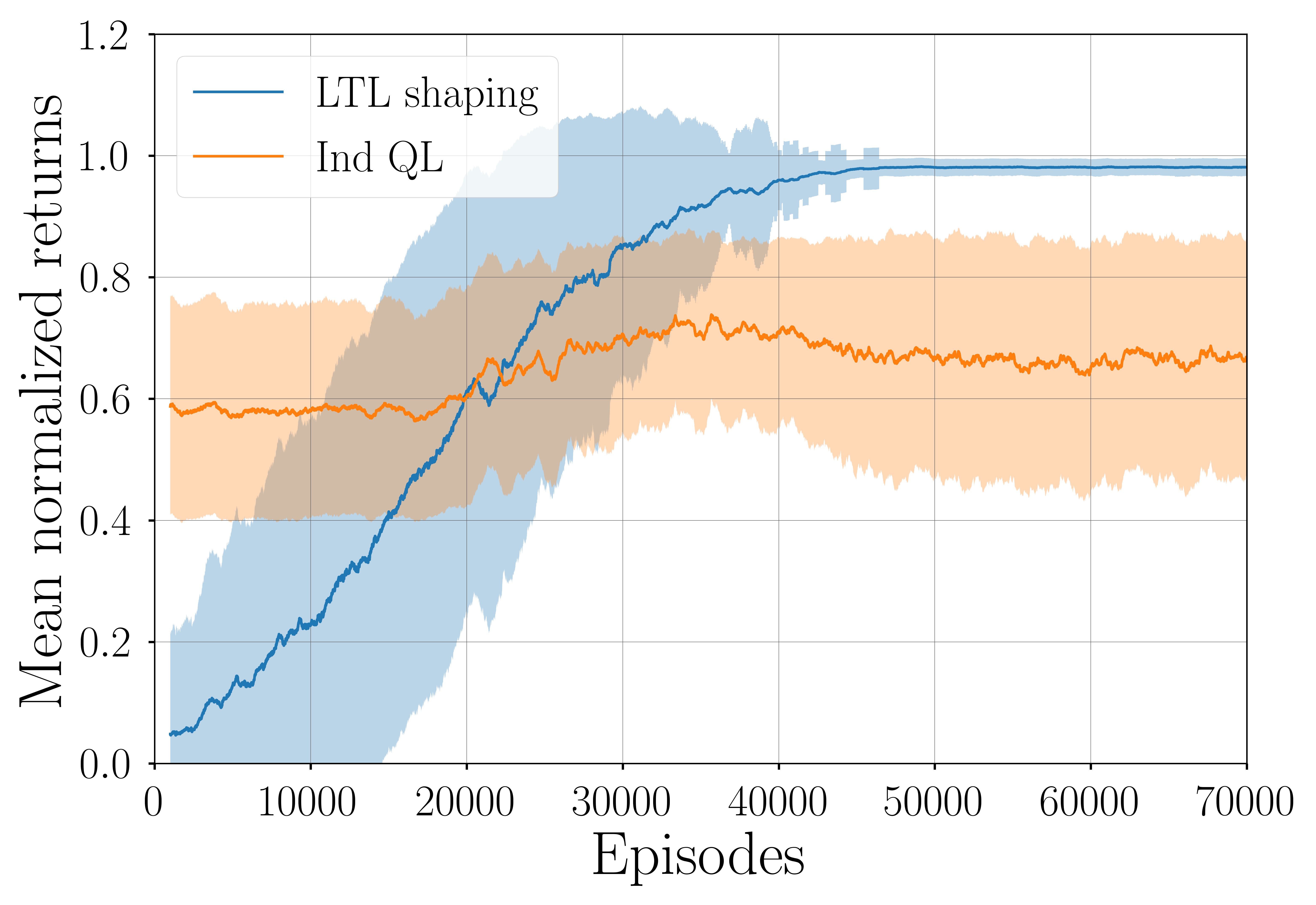}
  \label{fig:bench2_learning_curve}
}%
\caption{Results for the Flag Collection Example}
\label{fig:bench2}
\end{figure}

\begin{figure}
\centering
\subfloat [\emph{a}][Representation of the Environment MDP]{
  \raisebox{15 pt}{\includegraphics[width=.35\linewidth]{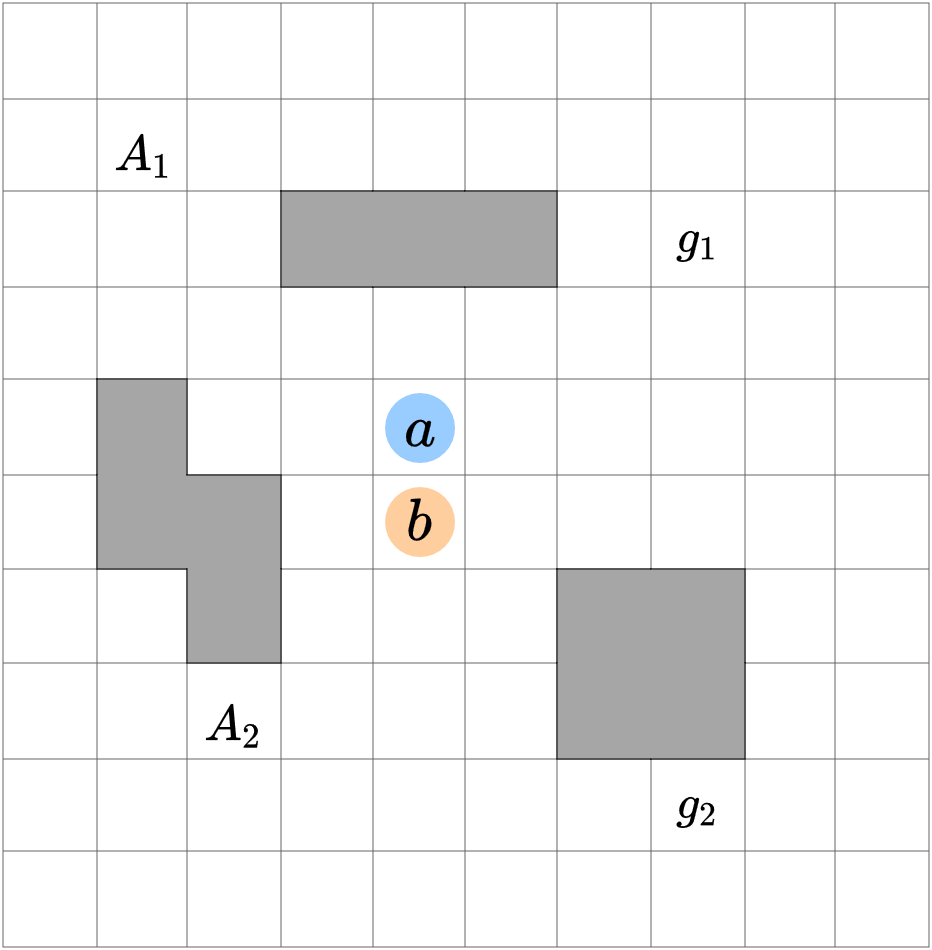}}
  \label{fig:bench3_mdp}
}
\subfloat [\emph{b}][Normalized Mean Return Learning Curve]{
  \includegraphics[width=.55\linewidth]{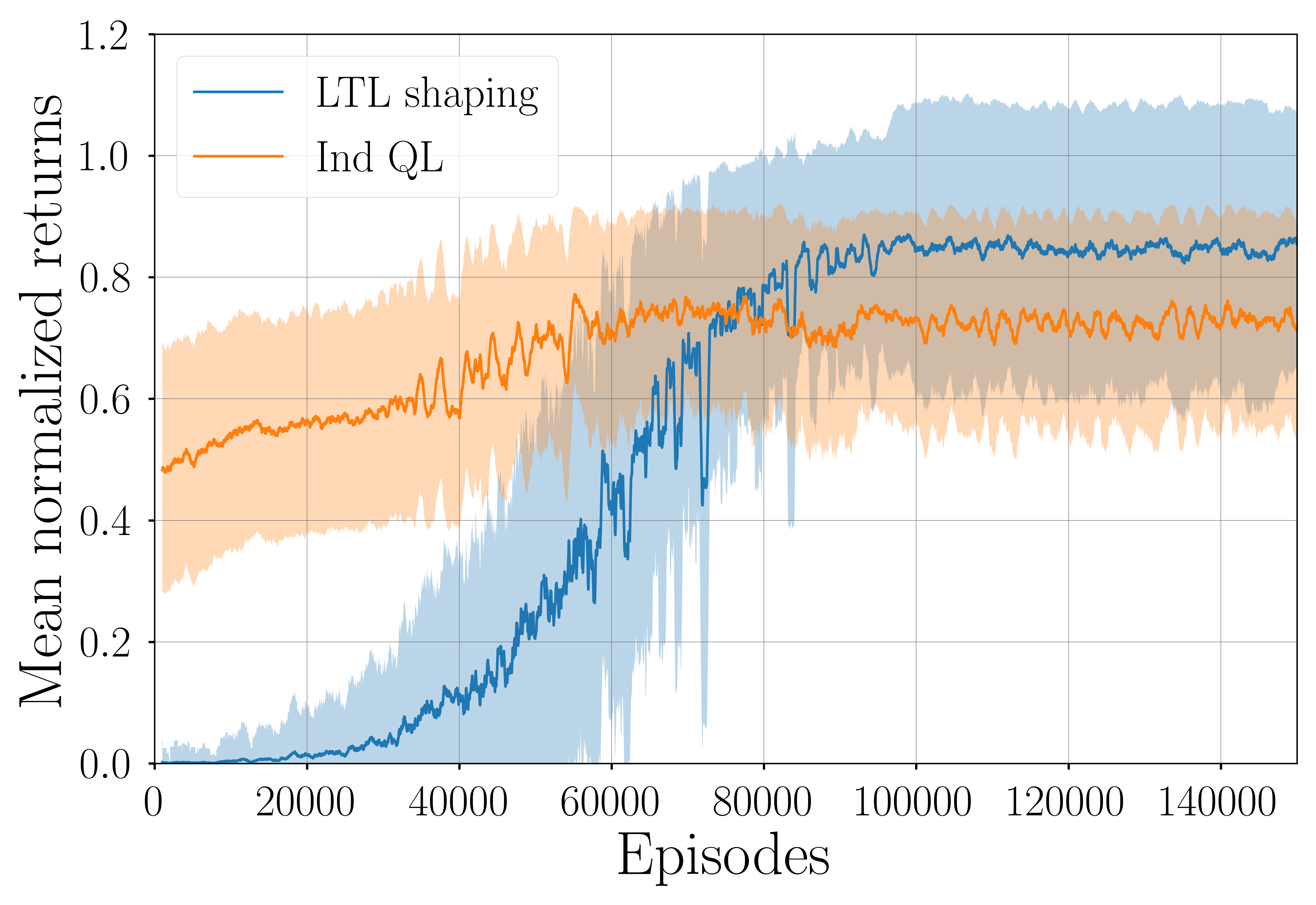}
  \label{fig:bench3_learning_curve}
}%
\caption{Rendez-vous}
\label{fig:bench3}
\end{figure}

Our method is implemented in Python \footnote{The code is available at : \url{https://github.com/IngyN/macsrl}} and is built upon the single agent RL tool (CSRL) developed in ~\cite{bozkurt2020control} which uses the Owl library to synthesize the LDBA from the LTL specifications ~\cite{kvretinsky2018owl}. We also use the Spot library ~\cite{spotlibrary} to visualize the automata synthesized for each example in Appendix \ref{spot_automata}. We applied our approach to three benchmarks including the previously discussed motivating example. The experiments were run on Windows Subsystem for Linux (WSL2) running Ubuntu with an AMD Ryzen 7 CPU with 32 GB of RAM. All experiments are averaged over the agents then smoothed using a rolling window and were each completed within one hour. We have not noticed a significant increase in time when running our method vs the same algorithm with no reward shaping. The main time complexity added by our method is the synthesis of the LDBAs and/or computation of the product MDP both of which only need to be computed once. For evaluation, we selected Independent Q-Learning ~\cite{bucsoniu2010multi} as a baseline but also as the underlying MARL algorithm for our method. Specifically, we used an $\epsilon$-greedy policy with $\gamma=0.999$ and $\gamma_B=0.99$ (accepting transitions discount). Both the probability for exploration and the learning rate are gradually decreased over the training from $1.0$ to ($0.01$ and $0.001$) respectively. For presented benchmarks, the environments visualized in figures \ref{fig:bench1_mdp}, \ref{fig:bench2_mdp} and \ref{fig:bench3_mdp} are non deterministic with a probability $p=0.8$ of going to the desired location and a probability $(1-p) = 0.2$ of ending up in a random other adjacent location. In all benchmarks, if an agent reaches a goal location (i.e. containing $g$ in the label), they cannot move for the remainder of the episode.  In order to compared the learning
performance we normalized the returns such that the minimum reward received corresponds to $0$ and the maximum corresponds to $1$ for each method. Moreover, for all graphs, we average the normalized returns over the agents and apply smoothing using a rolling window of $1000$ episodes.

 

\paragraph{Motivating Example.}Our first benchmark is the motivating example described previously in section \ref{motivating}. For the Independent Q-Learning baseline, we setup the independent reward function as follows: receive a reward of $2$ if both agent are at $a$ and $b$ at the same time, $10$ if they reach their goal. Notice that baseline reward function doesn't directly encourage agents pressing the buttons before reaching the goal. In contrast, with out approach, the agents do not receive a reward for reaching the goal locations until both buttons have been pressed simultaneously. The results observed (fig.\ref{fig:bench1_learning_curve}) show that our method with two different possible LTL specifications learns more reliably than the baseline with no reward shaping. In Section \ref{approach}, we discuss two possible specifications featuring different automata structures. The LTL shaping curve in blue corresponds to our initial specification $ \phi_3 = ((\Diamond a \land \neg g_1) \cup (a \land b)) \land \& ((\Diamond b \land \neg g_2) \cup (a\land b)) \land \& (\Diamond\square g_1) \land (\Diamond\square g_2)$. The LTL Shaping v2 curve in green corresponds to the specification $\phi_3'= \Diamond( (a\land b) \land \bigcirc\Diamond ( (g_1 \lor \bigcirc\Diamond g_1) \land ( g_2 \lor \bigcirc\Diamond g_2)) )$. The standard deviation over the rolling window depicted by the shaded area for $\phi_3'$ shows that the standard deviation is larger than the Independent Q-Learning baseline method. We hypothesize that this is because for the LDBA generated from the corresponding specification $\phi$ there only exists one accepting transition highlighted by a blue dashed line in figure \ref{fig:automatonf3}. This results in the agents either receiving a normalized reward of $1$ or $0$ in an episode which explains the larger standard deviation. the learning curve for the reward shaping based on $\phi_3'$ shows both better normalized returns per episode and better convergence. This difference in learning performance demonstrates the importance of choosing an adequate formats for LTL tasks. 

\paragraph{Flag collection.}Our second benchmark is a flag collection scenario inspired by ~\cite{neary2020reward}. Previous work has demonstrated this kind of scenario to be challenging for RL agents even in the single agent case ~\cite{elbarbari2021ltlf}. In this scenario (fig.\ref{fig:bench2_mdp}), the agents must collect flags $a$ and $b$ then proceed to a goal location ($g_1$, $g_2$). Similarly to the previous example, in the baseline case agents receive a reward of $2$ for collecting a flag and a reward of $10$ for reaching a goal. In this benchmark, no instruction or constraints are given as to which agent should go to which goal or collect which flags. The LTL specification used for this example is the following: $\phi = \Diamond (a \land \Diamond(b \land ( \Diamond( g_1 \lor \bigcirc\Diamond g_1) \land  \Diamond( g_2 \lor \bigcirc\Diamond g_2)) ))$. Using the $(label \lor \bigcirc\Diamond label)$ format helps create a larger number of accepting transitions (6 accepting transitions, with an automaton with 7 states) guiding the reward shaping progress better (figure \ref{fig:automaton_spot_2} in Appendix \ref{spot_automata}). In figure \ref{fig:bench2_learning_curve}, our method performs much better than the baseline which fails to converge with the same number of training episodes. In our testing, increasing the number of training episodes has not yielded better convergence for the Independent Q-Learning baseline. Notice that unlike the previous benchmark, our method has a much smaller standard deviation after convergence. 

\paragraph{Rendez-vous.}Our third benchmark is a rendez-vous task where agents must meet in the adjacent locations $a$ and $b$ then proceed to goal locations $g_1$ and $g_2$ (fig.~\ref{fig:bench3_mdp}). Similarly to the first benchmark, baseline agents receive a reward of $2$ if both agents are at locations $a$ and $b$ at the same time step and $10$ when each one reaches a goal location. In this scenario, agents are not assigned to a specific goal or meeting location, it is up to the learning algorithm to learn the optimal assignment. The LTL specification used for this benchmark is: $\phi =\Diamond( (a\land b) \land \bigcirc\Diamond ( (g_1 \lor \bigcirc\Diamond g_1) \land ( g_2 \lor \bigcirc\Diamond g_2)) )$. The synthesized LDBA is composed of $7$ states including the trap state and contains $5$ accepting transitions (fig.~\ref{fig:automaton_spot_3} in Appendix~\ref{spot_automata}). Figure \ref{fig:bench3_learning_curve} shows that in this slightly larger benchmark both methods converge but both still have a fairly large standard deviation with the average return for the LTL shaping method being higher than the baseline method with shaping.

\section{Discussion}
\label{discussion}
In order to assign tasks to specific agents, we initially attempted to use separate automatons that would communicate the perceived labels. In the motivating example (fig.~\ref{fig:motivatingex}), the first agent's automaton would check for button $b$ is triggered to be able to pass towards goal $g_1$. The second agent's automaton would include button $a$ that can only be reached by the first agent. However, without explicitly adding the labels from the other agents in the automaton, the product MDP for any one agent is incomplete and cannot reach the accepting states. In our motivating example from fig.~\ref{fig:motivatingex}, this would also happen if agent $A_2$'s LTL task was: $\phi_2= (\neg g_2 \cup a) \land \Diamond \square g_2 $.  Then the second agent could wait until label $a$ is triggered and immediately proceed to the goal in which case agent $A_1$ can never reach its accepting state. 

In the case of Independent Q-Learning used as the algorithm with our method, we argue that the augmented state $\langle s^i_t, q^i_t\rangle$ allows for indirect coordination by abstracting task progression more effectively. We can argue that if the joint LTL specification includes some task $A$ and some task $B$, when task $A$ is completed by an agent $i$ then the automaton state changes, implicitly communicating to the other agents that task $A$ is complete. 

In this method, we assume that if agents choose conflicting $\epsilon$-action then the first agent's choice in the agent ordering decides which one is being taken thereby enforcing a priority scheme based on agent ordering. For example, if we are using the LDBA from figure \ref{fig:ldba} one agent may choose to take the $\epsilon$-action leading to $q_1$ and the other the $\epsilon$-action leading to $q_2$, which transition should be taken? The more agents, we consider, the more likely conflicting $\epsilon$-actions will be chosen. This is currently one of our main limitations and are exploring how the use of a more dynamic agent priority scheme or negotiation could be used to improve this.

The difference in performance in our experiments have shown that to have optimal reward shaping, we may need to formulate the specification in such a way to increase the number of accepting transitions in the LDBA. Moreover, different methods for reward shaping create the reward function based on the automaton very differently. For example, in the approach presented in ~\cite{elbarbari2021ltlf}, the authors use heuristics and attempt to estimate the progress within the LDBA to better guide the learning process. 

For the experiments section, we also attempted to demonstrate that this method could work with another algorithm (we tried with shared-experience Q-learning ~\cite{christianos2020shared}) in a slightly modified foraging gym environment ~\cite{papoudakis2020comparative}. The agents were unable to learn in our attempts. However, we believe it was an issue with the implementation of the necessary modifications rather than a failure of our method. 

\section{Conclusion}
\label{conclusion}
In this work, we present an approach to multi-agent reward shaping that can successfully learn LTL defined tasks. The semi-centralized reward shaping approach addresses the issue of scalability and the need for coordination between agents. We also demonstrate how the use of LTL in this approach can allow for flexible task specification and assignment. In our experimental results, we highlight that our method can achieve encouraging learning performance when compared to Independent Q-Learning alone. In the future, we may look into ways to improve the logic-based reward shaping mechanism further by exploring how this method would perform in the context of lossy or restricted communication. It is our belief that a slightly modified main approach would be a useful tool in this context. Another idea would be to use a hierarchical approach for team based tasks. Furthermore, we may seek to revisit the decentralized approach and analyze how it can be achieved in a modified problem. Finally, we may explore the use of our method with a better way of determining progress through the automaton rather than waiting until agents reach accepting transitions. 

\section{Acknowledgements}
We thank Haiying Shen, Haifeng Xu and Hongning Wang on their advice and feedback for this project. We also thank Alper Kamil Bozkurt for his help and collaboration.




\newpage
\appendix
\section{Experiment Automata}
\label{spot_automata}
\begin{figure}[h]
\centering
\includegraphics[scale=0.4]{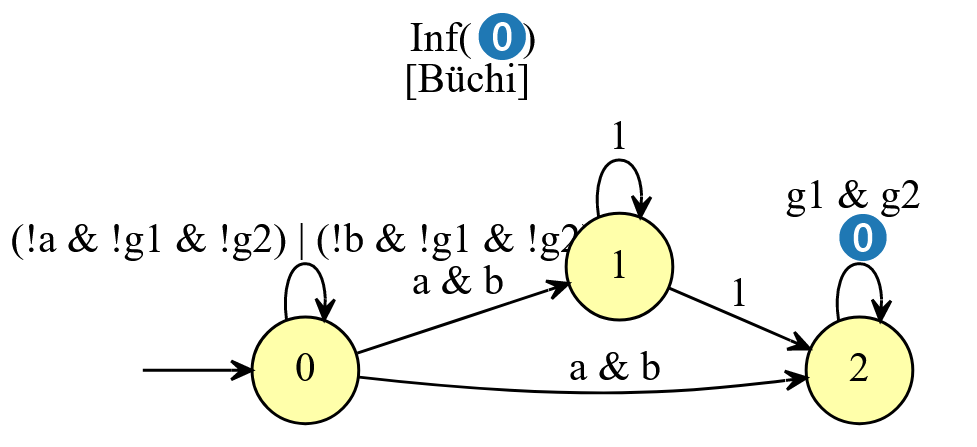}
\caption{Spot Generated Automaton For Original Motivating Example LTL $\phi_3$}
\label{fig:automaton_spot_1}
\end{figure}

In this appendix, we present the automata synthesized for each benchmark in the Experiments (section \ref{experiments}). The transitions annotated with a blue $0$ are the accepting transitions. The automata seen here do not show the trap state to which agents would transition in the event of traces violating the specification. For example, in state $0$ for the automaton in figure \ref{fig:automaton_spot_1}, if either agent went to a goal location ($g1$ or $g2$) that would violate the specification and the common automaton would transition to the trap state.  

\begin{figure}[h]
\centering
\includegraphics[scale=0.45]{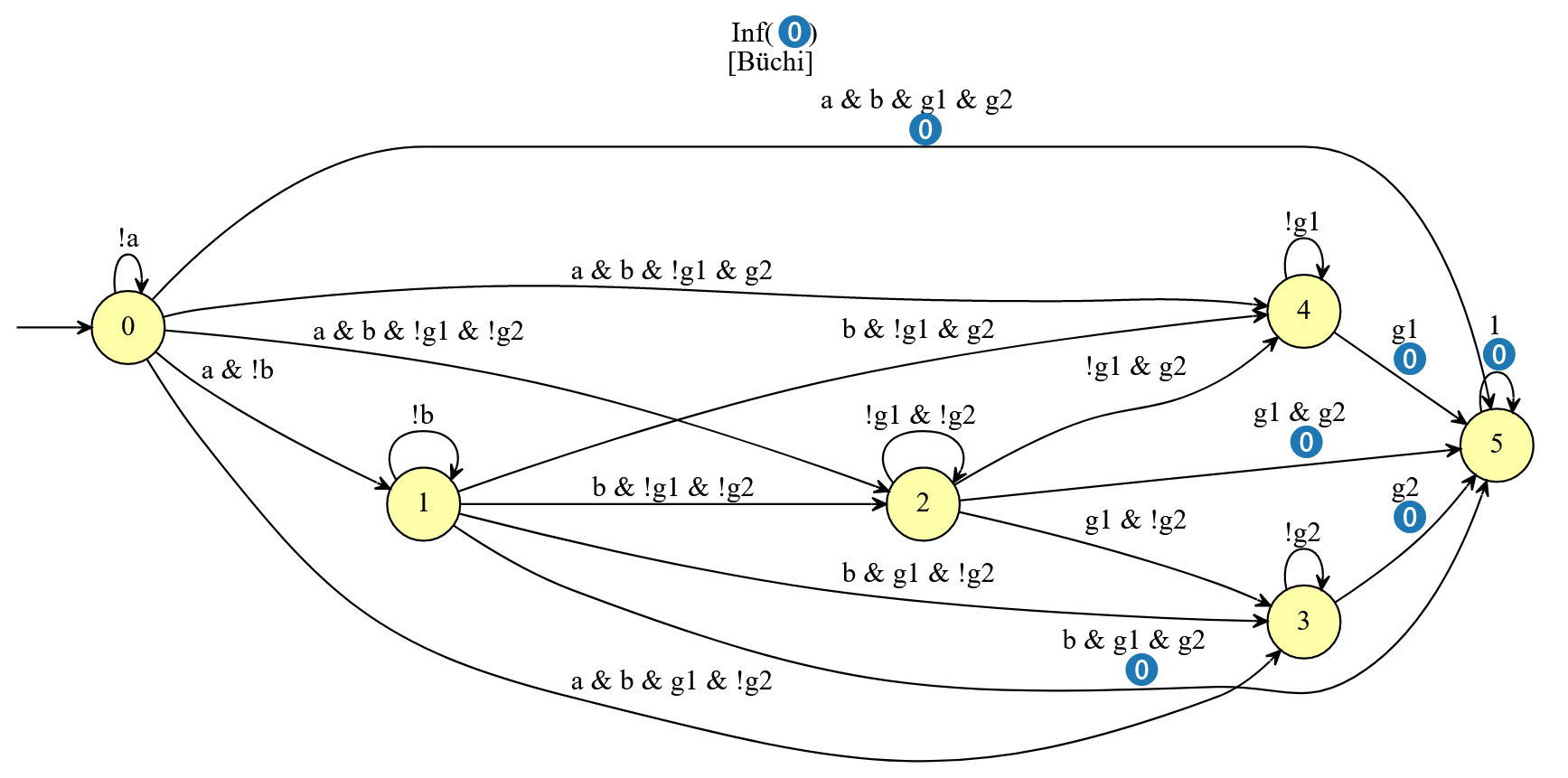}
\caption{Spot Generated Automaton For Flag Collection Scenario}
\label{fig:automaton_spot_2}
\end{figure}

\begin{figure}[ht]
\centering

\includegraphics[scale=0.53]{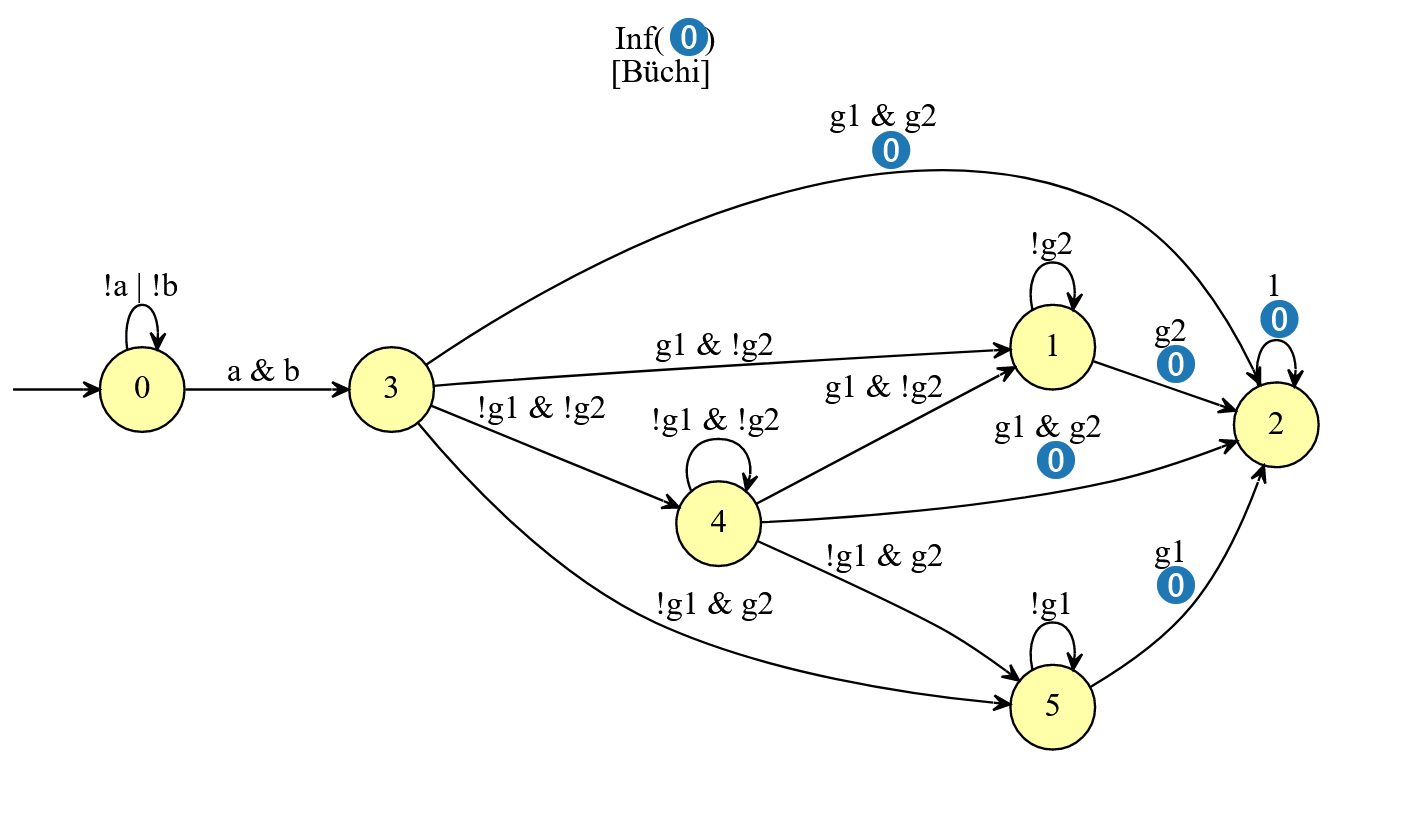}
\caption{Spot Generated Automaton For Rendez-vous Scenario and for Motivating example $\phi_3'$}
\label{fig:automaton_spot_3}
\end{figure}

\end{document}

%% file: algorithm1.tex
\begin{algorithm}[t]
\caption{Semi-centralized reward shaping at time step $t$ for an agent $i$} \label{algorithm1}
\small
\begin{algorithmic} [1]
\Require Environment $\mathcal{M}_i$ , MARL algorithm $\mathbb{A}$, automaton $\mathcal{A}^\phi$ and augmented state $\langle s_t^i, q_t \rangle$

\State $\epsilon_t = \emptyset$
\ForAll{agent $i$ }
	\State $a_t^i = \mathbb{A}.\texttt{action\_selection}(\langle s_t^i, q_t \rangle,\mathcal{A}^i_t)$
	\Comment{ Check for epsilon actions}
    \If {$a^i_t \in \mathcal{E}$ }
    	\State $\epsilon_t.append(a^i_t)$
    \EndIf 
    \State $s^i_{t+1} = \mathcal{M}_i.\texttt{environment\_step}(s^i_t \setminus\{q_t\}, a^i_t \setminus \{\epsilon_t\})$

\EndFor

\State $l_t = get\_global\_labels(s_{t+1})$ 
\If{$\epsilon_t \neq \emptyset$}
	\State $q_{t+1}, r_t =\mathcal{A}^\phi. \texttt{update\_automaton}(q_t, l_t, \epsilon_t)$ \Comment{Update automaton state and retrieve reward}
\EndIf

\State $\mathbb{A}.\texttt{update}(\langle s_t, q_t\rangle ,  \langle s_{t+1},q_{t+1}\rangle, r_t, a_t )$

\end{algorithmic}
\end{algorithm}